\documentclass[sigconf,natbib=true,anonymous=false,screen=true,review=false]{acmart}

\usepackage{color}
\usepackage{graphicx}
\usepackage{acronym}
\usepackage{CJK}
\usepackage{booktabs}
\usepackage[inline]{enumitem}
\usepackage[skip=0pt]{caption}
\usepackage{amsmath}
\usepackage{booktabs}
\usepackage[ruled,linesnumbered,noend]{algorithm2e}
\usepackage{multirow}
\definecolor{darkpastelgreen}{rgb}{0.01, 0.75, 0.24}
\usepackage{bbm}

\acrodef{MDS}{medical dialogue system}
\acrodef{TDS}{task-oriented dialogue system}
\acrodef{ODS}{open-ended dialogue system}
\acrodef{NLU}{natural language understanding}
\acrodef{DPL}{dialogue policy learning}
\acrodef{DST}{dialogue state tracking}
\acrodef{NLG}{natural language generation}
\acrodef{DM}{dialogue management}
\acrodef{SeqMDS}{sequential medical dialogue system}
\acrodef{$M^2$-MedDialog}{multiple-domain multiple-service medical dialogue}
\acrodef{RL}{reinforcement learning}
\acrodef{SL}{semantic label}
\acrodef{CL}{contrastive learning}
\acrodef{SCL}{self-supervised contrastive learning}
\acrodef{NLP}{natural language processing}

\acrodef{ReMeDi}{resources for medical dialogues}
\newcommand{\OurResources}{\ac{ReMeDi}}
\copyrightyear{2022}
\acmYear{2022}
\setcopyright{rightsretained}
\acmConference[SIGIR '22]{Proceedings of the 45th International ACM SIGIR Conference on Research and Development in Information Retrieval}{July 11--15, 2022}{Madrid, Spain}
\acmBooktitle{Proceedings of the 45th International ACM SIGIR Conference on Research and Development in Information Retrieval (SIGIR '22), July 11--15, 2022, Madrid, Spain}
\acmDOI{}
\acmISBN{}

\author{%
Guojun Yan\textsuperscript{\rm 1}
\quad Jiahuan Pei\textsuperscript{\rm 2}
\quad Pengjie Ren\textsuperscript{\rm 1}
\quad Zhaochun Ren\textsuperscript{\rm 1}
\quad Xin Xin\textsuperscript{\rm 1} 
\\ 
Huasheng Liang\textsuperscript{\rm 3}
\quad Maarten de Rijke\textsuperscript{\rm 2}
\quad Zhumin Chen\textsuperscript{\rm 1}
}

\affiliation{
    \textsuperscript{\rm 1}Shandong University, Qingdao \country{China}\\
    \textsuperscript{\rm 2}University of Amsterdam, Amsterdam \country{The Netherlands}\\
    \textsuperscript{\rm 3}WeChat Tencent, Shenzhen \country{China}\\
    yan\_gi@mail.sdu.edu.cn,
    \{renpengjie, zhaochun.ren, xinxin, chenzhumin\}@sdu.edu.cn, \\ \{j.pei, m.derijke\}@uva.nl, watsonliang@tencent.com
}

\def\authors{Guojun Yan, Jiahuan Pei, Pengjie Ren, Zhaochun Ren, Xin Xin, Huasheng Liang, Maarten de Rijke and Zhumin Chen}

\renewcommand{\shortauthors}{Yan et al.}

\title[ReMeDi: Resources for Multi-domain, Multi-service, Medical Dialogues]{ReMeDi: Resources for \\ Multi-domain, Multi-service, Medical Dialogues}

\date{}

\begin{document}


\begin{abstract}
\Acp{MDS} aim to assist doctors and patients with a range of professional medical services, i.e., diagnosis, treatment and consultation. 
The development of \acp{MDS} is hindered because of a lack of resources. 
In particular.
\begin{enumerate*}[label={(\arabic*)}] 
\item there is no dataset with large-scale medical dialogues that covers multiple medical services and contains fine-grained medical labels (i.e., intents, actions, slots, values), and 
\item there is no set of established benchmarks for \acp{MDS} for multi-domain, multi-service medical dialogues.
\end{enumerate*}

In this paper, we present \acs{ReMeDi}, a set of \acl{ReMeDi}\acused{ReMeDi}.
\OurResources{} consists of two parts, the \OurResources{} dataset and the \OurResources{} benchmarks.
The \OurResources{} dataset contains 96,965 conversations between doctors and patients, including 1,557 conversations with fine-gained labels.
It covers 843 types of diseases, 5,228 medical entities, and 3 specialties of medical services across 40 domains.
To the best of our knowledge, the \OurResources{} dataset is the only medical dialogue dataset that covers multiple domains and  services, and has fine-grained medical labels.

The second part of the \OurResources{} resources consists of a set of state-of-the-art models for (medical) dialogue generation.
The \OurResources{} benchmark has the following methods:
\begin{enumerate*}
\item pretrained models (i.e., BERT-WWM, BERT-MED, GPT2, and MT5) trained, validated, and tested on the \OurResources{} dataset, and 
\item a \acf{SCL} method to expand the \OurResources{} dataset and enhance the training of the state-of-the-art pretrained models.
\end{enumerate*}

We describe the creation of the \OurResources{} dataset, the \OurResources{} benchmarking methods, and establish experimental results using the \OurResources{} benchmarking methods on the \OurResources{} dataset for future research to compare against.
With this paper, we share the dataset, implementations of the benchmarks, and evaluation scripts.
\end{abstract}



\maketitle

\acresetall


\section{Introduction}
\label{01-introduction}

Medical research with AI-based techniques is growing rapidly~\cite{zhang2021smedbert,kao2018context,chintagunta2021medically,wang2020coding}.
\Acp{MDS} promise to increase access to healthcare services and
to reduce medical costs~\cite{zeng2020meddialog,yang2020generation,li2021semi}.
\Acp{MDS} are more challenging than common \acp{TDS} for, e.g., ticket or restaurant booking~\cite{li2017end,peng2018adversarial,wen2017network} in that they require a great deal of expertise.
For instance, there are much more professional terms, which are often expressed in colloquial language~\cite{shi2020understanding}.

\begin{figure}[!t]
    \centering
    \includegraphics[trim = 0.5mm 1.0mm 0.1mm 0.1mm, clip, width=0.47\textwidth]{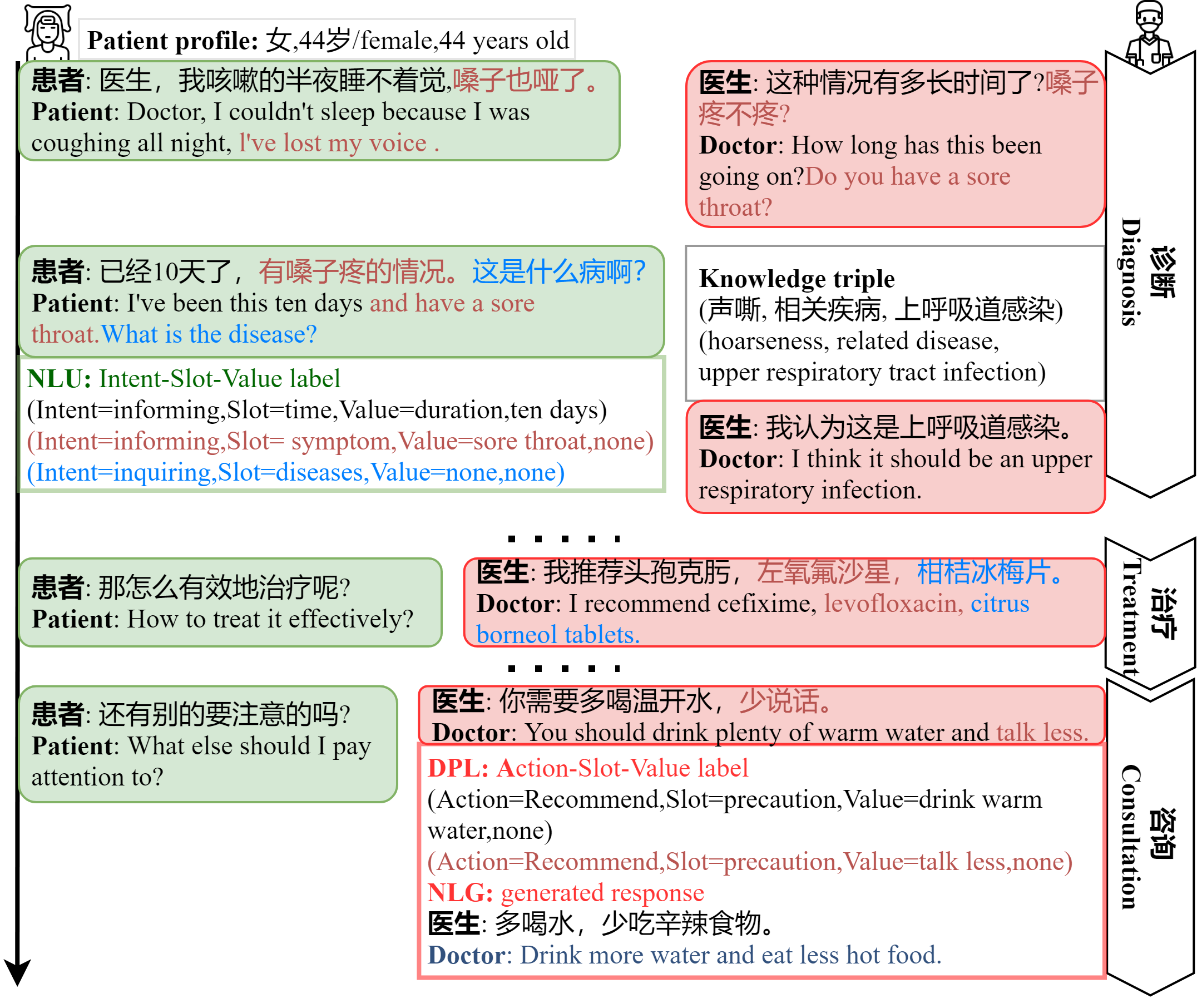}
    \caption{A practical medical dialogue involving diagnosis,  treatment and consultation. They are all dependent. Combined with the knowledge triple in the upper right corner, we can better infer the related diseases. The lower right part is our annotation example, including intent/action, slot and value.
    }
    \label{fig:example}
\end{figure}

Recently, extensive efforts have been made towards building data for \acs{MDS} research~\cite{liao2020task,shi2020understanding}.
Despite these important advances, limitations persist:
\begin{enumerate*}
\item In currently available datasets, there is a lack of a complete diagnosis and treatment procedure.
A practical medical dialogue is usually a combination of consultation, diagnosis and treatment, as shown in Figure~\ref{fig:example}.
To the best of our knowledge, no previous study considers all three medical services simultaneously~\citep{wei2018task,xu2019end,liu2020meddg,yang2020generation}.
\item In currently available datasets, labels are not comprehensive enough.
Most datasets only provide the slot-value pairs for each utterance.
Intent labels and medical knowledge triples related to each utterance are rarely provided.
For example, there is one utterance in~\citep{zhang2020mie}: ``Patient: Doctor, could you please tell me is it premature beat?'' 
It only has the slot-value label ``Symptom: Cardiopalmus'', without the intent label ``Inquire'' and the required knowledge triple ``<premature beat, symptom, cardiopalmus>''.
\item In currently available datasets, labels are not fine-grained enough.
Composite utterances, which contain more than one intent/action, are common in practice.
For example, for the third utterance in Figure~\ref{fig:example}, the patient says ``Ten days. Yes. What is the disease?'', there are three kinds of intents: informing time, informing symptom status, and inquiring diseases.
Previous studies usually provide a single coarse-grained label for the whole composite utterance, which might mislead the training of models and/or lead to inaccurate evaluation.
Second, we find that the values defined in previous work can hardly accurately convey complex information.
Instead, we provide main-subordinate values, each of which includes a main value and a subordinate value.
For example, for the labeling ``Value=duration, ten days'' of the second user utterance in Figure~\ref{fig:example}, the main value is ``duration'' and the subordinate value is ``ten days''. 
The main-subordinate values have a stronger capacity to convey complex information:
\begin{enumerate*}
\item Negation status of an entity, e.g., without experiencing symptom sore throat.
\item The specific value of an entity, e.g., the specific number of blood pressure.
\item Relationship between entities, e.g., the side effect of a medicine. 
\end{enumerate*}
\item Besides the limitations above,
some datasets only involve limited medical entities.
For example, MedDG~\cite{zeng2020meddialog}, a very recent medical dialogue dataset, only contains 12 diseases.
\end{enumerate*}

To address the lack of a suitable dataset, our first contribution in this paper is the introduction of the \acfi{ReMeDi} dataset. 
The \OurResources{} dataset has the following features:
\begin{enumerate*}
\item medical dialogues for consultation, diagnosis and treatment, as well as their mixture;
\item comprehensive and fine-grained labels, e.g., intent-slot-value triples for sub-utterances; and
\item more than 843 diseases, 20 slots and 5,228 medical entities are covered.
\end{enumerate*}
Moreover, we ground the dialogues with medical knowledge triples by mapping utterances to medical entities.

Our second contribution in this paper is a set of medical dialogue models for benchmarking against the \OurResources{} dataset.
Recent work considers \acp{MDS} as a kind of \ac{TDS}~\cite{wei2018task,xu2019end,liao2020task} by decomposing a \acs{MDS} system into well-known sub-tasks, e.g., \ac{NLU}~\citep{shi2020understanding}, \ac{DPL}~\citep{wei2018task}, and \ac{NLG}.
There is, however, no comprehensive analysis on the performance of all the above tasks when achieved and/or evaluated simultaneously.
To establish a shared benchmark that addresses all three \ac{NLU}, \ac{DPL}, and \ac{NLG} tasks in a \ac{MDS} setting, we adopt causal language modeling, use several pre-trained language models (i.e., BER-WWM, BERT-MED, MT5 and GPT2) and fine-tune them with the \OurResources{} dataset. 
In addition, we provide a pseudo labeling algorithm and a natural perturbation method to expand the proposed dataset, and enhance the training of state-of-the-art pretrained models based on \acl{SCL}.

In the remainder of the paper, we detail the construction of the \OurResources{} dataset and the definition of the \OurResources{} benchmarks, and evaluate the \OurResources{} benchmarks against the \OurResources{} dataset on the \ac{NLU}, \ac{DPL}, and \ac{NLG} tasks, thereby establishing a rich set of resources for \acl{MDS} to facilitate future research. 
Details on obtaining the resources are included in the appendix of this paper.


\section{Related Work}
We survey related work in terms of datasets, models and contrastive learning.

\subsection{Medical dialogue datasets}
Most medical dialogue datasets contain only one domain, e.g., Pediatrics~\cite{wei2018task,xu2019end,liao2020task}, COVID-19~\cite{yang2020generation}, Cardiology~\cite{zhang2020mie}, Gastroenterology~\cite{liu2020meddg} and/or one medical service, e.g., Diagnosis~\cite{lin2019enhancing,lin2021graph}, Consultation~\cite{shi2020understanding}.
However, context information from other services and/or domains is often overlooked in a complete medical aid procedure.
For example, in Figure ~\ref{fig:example}, the symptom ``sore throat'' mentioned in the diagnosis service has the long-term effect on the suggestion ``talk less'' in the follow-up consultation service.
To this end, we provide medical dialogues for consultation, diagnosis and treatment, as well as their mixture in the \OurResources dataset. 
Although a few datasets~\cite{zeng2020meddialog,li2021semi} contain multiple medical services in multiple domains, they target the \acs{NLG} only without considering the \ac{NLU} and \ac{DPL}.
Differently, \OurResources contains necessary labels for \ac{NLU}, \ac{DPL} and \acs{NLG}.
Another challenge of existing datasets is the medical label insufficiency problem.
The majority of datasets only provide a spot of medical labels for slots or actions, e.g., one slot~\cite{shi2020understanding,du2019extract,lin2019enhancing}, single-value~\cite{liu2020meddg,lin2021graph,li2021semi}.
Moreover, their labels are too coarse to distinguish multiple intents or actions in one utterance.
Unlike all datasets above, our dataset provides comprehensive and fine-grained intent/action labels for constituents of an utterance.

To sum up, \OurResources{} is the first multiple-domain multiple-service medical dialogue dataset with fine-grained medical labels and large-scale entities, which is more competitive compared with the datasets mentioned above in terms of 9 aspects (i.e., domain, service, task, intent, slot, action, entity, disease, dialogue).
A summary can be found in Table~\ref{com}.

\begin{table*}[!htb]
\setlength{\tabcolsep}{4.8pt}
\centering
\caption{Comparison between the proposed corpora and other medical dialogue corpora.}
\label{com}
		\begin{tabular}{llllcrrrrr}			
		\toprule
			\bf Dataset & \bf (\#)Domains & \bf (\#)Services & \bf (\#)Tasks & \bf \#Intents/Slots/Actions & \bf \#Entities  & \bf\#Diseases & \bf\#Dialogues \\
			\midrule
			CMDD \cite{lin2019enhancing} & Pediatrics & Diagnosis & \ac{NLU} & - / 1 / - & 162  & 4 & 2,067\\
			SAT \cite{du2019extract} & 14 & \textbf{3} & \ac{NLU} & - / 1 / - & 186 & - & 2,950\\
			MSL \cite{shi2020understanding}  & Pediatrics & Consultation & NLU & - / 1 / - & 29  & 5 & 1,652\\
			MIE \cite{zhang2020mie} & Cardiology & 2 & \ac{NLU} & - / 4 / - & 71  & 6& 1,120 \\
		    MZ \cite{wei2018task} & Pediatrics & Diagnosis & DPL & - / 2 / 6 &67 & 4  & 710 \\ 
			DX \cite{xu2019end} & Pediatrics & Diagnosis & DPL &- / 2 / 5 & 41 & 5 & 527 \\
			RD \cite{liao2020task} & Pediatrics & Diagnosis & DPL & - / 2 / 2 & 90 & 4 & 1,490 \\
			SD \cite{liao2020task} & 9 & Diagnosis & DPL & - / 2 / 2 & 266  & 90 & 30,000\\
			COVID-EN \cite{yang2020generation} & COVID-19 & \textbf{3} & NLG & - / - / -& - & 1 & 603 \\
			COVID-CN \cite{yang2020generation} & COVID-19 & \textbf{3} & NLG & - / - / - & -  & 1 & 1,088\\
			MedDG \cite{liu2020meddg} & Gastroenterology & 2 & NLG & - / 5 / - & 160  & 12 & 17,864\\
			MedDialog-EN \cite{zeng2020meddialog} & 51 & \textbf{3} & NLG & - / - / - & -  & 96 & 257,332\\
			MedDialog-CN \cite{zeng2020meddialog} &29 & \textbf{3} & NLG & - / - / - & -  & 172 & \textbf{3,407,494}\\
			Chunyu \cite{lin2021graph} & - & Diagnosis & NLG & - / 2 / - & -  & 15 & 12,842\\
			KaMed \cite{li2021semi} & \textbf{100} & \textbf{3} &  NLG & - / 4 / - & \textbf{5,682}  & -& 63,754\\ 
			\midrule
		    \OurResources{}-base & 30 & \textbf{3} & \textbf{3} & \textbf{5}/\textbf{20}/\textbf{7}  & 4,825   & 491& 1,557\\
		    \OurResources{}-large & 40 & \textbf{3} & \textbf{3} & \textbf{5}/\textbf{20}/\textbf{7} & 5,228   & \textbf{843}& 95,408\\
		\bottomrule
		\end{tabular}	
\end{table*}

\subsection{Pretrained models for \aclp{TDS}}
Large language models have achieved state-of-the-art performance for \acp{TDS}~\cite{zhang2020recent,balaraman2021recent}. 
BERT~\cite{devlin2018bert} is widely used as a benchmark for \acp{TDS}~\cite{zhu2020crosswoz,mehri2020dialoglue} and has been shown to be effective for understanding and tracking dialogue states~\cite{lai2020simple, chao2019bert}.
In terms of dialogue generation, BERT is usually used in a selective way (e.g., TOD-BERT~\cite{wu2020tod}).
The GPT family of language models~\cite{radford2018improving,radford2019language} serves as a competitive and common benchmark in recent work on \acp{TDS}~\cite{budzianowskiV2019hello,hosseini2020simple,yang2020ubar}.
GPT is used as a promising backbone of recent research on generating dialogue states~\cite{wu2020tod,melas2019generation,chang2021jointly} and actions~\cite{li2021language}.
MT5~\cite{raffel2019exploring} is the current benchmark for \acp{TDS}, because it  
inherits T5~\cite{Raffel20T5}'s powerful capabilities of text generation and provides with multilingual settings~\cite{lin2021bitod,zuo2021allwoz,majewska2022cross}.
Large neural language models are data hungry and data acquisition for \acp{TDS} is expensive~\cite{razumovskaia2021crossing}.
An effective method for alleviating this issue is \ac{CL}.
\acs{CL} compares similarity and dissimilarity by positive/negative data sampling, and defining contrastive training objectives~\cite{le2020contrastive,khoslaTWSTIMLK20super}.
Most studies work on re-optimizing the representation space based on contrastive word~\cite{mrkvsic2016counter,pei2016combining,huang2018incorporating,GutmannH12nosie} or sentence~\cite{KirosZSZUTF15skip,jason19eda,fang2020cert,shen20asimple,gao21simcse} pairs.
Some also focus on sampling negative data pairs~\cite{xiong21approximate,leeLH21contras,wang2022sncse,wang21CLINE}.
Research has also explored different contrastive training objectives based on single~\cite{chopra2005learning,gutmann2010noise} or multiple~\cite{schroff2015facenet,sohn2016improved,chen20SIMCLR} positive and negative pairs, along with their complex relations~\cite{oh2016deep,salakhutdinov2007learning,frosst2019analyzing}.

In this work, we share several pretrained language models based on BERT, GPT2, MT5 as benchmarks of \OurResources{}.
To alleviate the data hungry problem, we enhance the pretrained language models with \acl{CL}.

\subsection{Medical dialogue systems}
Similar to \acp{TDS}~\cite{chen2017survey}, a \acs{MDS} system can be divided into several sub-tasks, e.g., \ac{NLU}, \ac{DPL}, and \ac{NLG}.

\ac{NLU} aims to understand user utterances by intent detection~\cite{wei2018task} and slots filling~\cite{weld2021NLU,chen2019WAIS,qin2019stack}. 
\citet{du2019extract,du2019learning} formulate \ac{NLU} as a sequence labeling task and use Bi-LSTM to capture contextual representation for filling entities and their relations into slots. 
\citet{lin2019enhancing} improve filling entities with global attention and symptom graph.
\citet{shi2020understanding} propose the label-embedding attentive multi-label classifier and improve the model by weak supervision from responses.
\ac{DST} tracks the change of user intent~\cite{mrkvsic2017neural}.
\citet{zhang2020mie} employ a deep matching network, which uses a matching-aggregate module to model turn-interaction among utterances encoded by Bi-LSTM.
In this work, we integrate \ac{DST} into vanilla \ac{NLU} to generate intents and updated slot values simultaneously. 

\ac{DPL} decides system actions given a set of slot-value dialogue states and/or a dialogue context~\cite{chen2017survey}.
\citet{wei2018task} first adopt \ac{RL} to extract symptoms as actions for disease diagnosis.
\citet{xu2019end} apply deep Q-network based on a medical knowledge graph to track topic transitions.
\citet{xia2020generative} improve \ac{RL} based \ac{DPL} using generative adversarial learning with regularized mutual information.
\citet{liao2020task} use a hierarchical \ac{RL} model to alleviate the large action space problem.
We generate system actions as general tokens to fully avoid action space exploration in these \ac{RL} models.

\ac{NLG} generates system responses given the outputs from \ac{NLU} and \ac{DPL}~\cite{pei2019modular}.
\citet{yang2020generation} apply several pretrained language models (i.e., Transformer, GPT, and BERT-GPT) to generate doctors' responses for COVID-19 medical services.
\citet{liu2020meddg} provide several \ac{NLG} baselines based on sequence-to-sequence models (i.e., Seq2Seq, HRED) and pretrained language models (i.e., GPT2 and DialoGPT).
\citet{li2021more} use pretrained language models to predict entities and generate responses.
Recently, meta-learning ~\cite{lin2021graph} and semi-supervised variational Bayesian inference \cite{li2021semi} are adopted for low-resource medical response generation.


\section{The \protect\OurResources{} Dataset}
\begin{figure}[htb]
   \centering
    \includegraphics[trim = 0mm 0mm 0.5mm 0.6mm, clip, width=0.46\textwidth, height=0.2\textwidth]{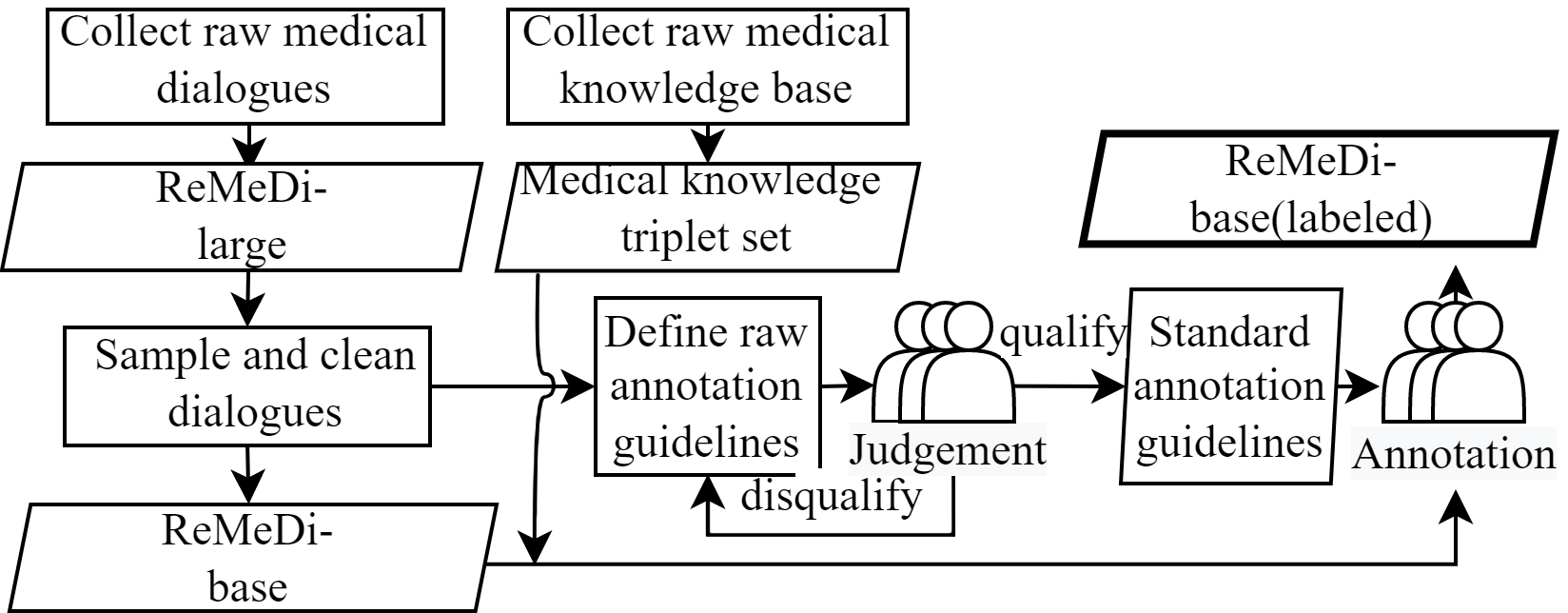}
    \caption{Process of the \OurResources{} dataset construction. }
    \label{process}
\end{figure}

The \OurResources{} dataset is built following the pipeline shown in Figure~\ref{process}:
\begin{enumerate*}[label={(\arabic*)}] 
\item We collect raw medical dialogues and knowledge base from online websites;
\item We clean dialogues by a set of reasonable rules, and sample dialogues by considering the proportions of disease categories;
\item We define annotation guidelines and incrementally improve them by dry-run annotation feedbacks until standard annotation guidelines are agreed by annotators;
\item We conduct human annotation with standard annotation guidelines.
\end{enumerate*}

Note that we provided two versions of the dataset: a labeled \OurResources{}-base (1,557 dialogues) and an unlabeled \OurResources{}-large (95,408 dialogues).
The former is for evaluating the performance of the benchmark models and the latter is for improving the training of large models (see details in \S\ref{sec:cl}).


\subsection{Collecting raw dialogues and knowledge base}
We collect 95,408 natural multiple-turn conversations between doctors and patients from ChunYuYiSheng,\footnote{\url{https://www.chunyuyisheng.com/}} a Chinese online medical community.
All information from the website is open to the public and has been processed with ethical considerations by the website, e.g., the sensitive information of patients, such as names, has been anonymized.
To further ensure data privacy, we anonymize more potentially sensitive information, e.g., the name of doctors, the address of hospitals, etc.
These raw dialogues constitute a large-scaled unlabeled dataset, called \OurResources{}-large.
It covers 40 domains (e.g., pediatrics), 3 services (i.e., diagnosis, consultation, and treatment), 843 diseases (e.g., upper respiratory tract infection), and 5,228 medical entities.
%
We crawled 2.6M medical $<$entity1, relation, entity2$>$ triplets from CMeKG2.0,\footnote{\url{http://cmekg.pcl.ac.cn/}} a Chinese medical knowledge base.
For example, the triplet $<$paracetamol, indication, headache$>$ denotes paracetamol can relieve headache.
The entities involve about 901 diseases, 920 drugs, 688 symptoms, and 200 diagnosis and treatment technologies. 
The number of relation types is 125.

\subsection{Cleaning and sampling dialogues}
We conduct the following steps to obtain a set of dialogues for human annotation:
\begin{enumerate*}[label={(\arabic*)}]   
\item 
Filtering out noise dialogues.
First, we filter out short-turn dialogues with less than 8 utterances, because we find these short dialogues usually do not contain much information.
Next, we filter out inaccurate dialogues with images or audios and keep dialogues with literal utterances only.
Finally, we filter out dialogues in which too few medical entities emerged in the crawled knowledge triplet set.
\item Anonymizing sensitive information.
We use special tokens to replace sensitive information in raw dialogues, e.g., ``[HOSPITAL]'' is used to anonymize the specific name of a hospital.
\item  Sampling dialogues by disease categories. 
In order to balance the distribution of diseases, we extract the same proportion of dialogues from each disease to form \OurResources{}-base for annotation.
\end{enumerate*}


\subsection{Incremental definition of annotation guidelines}
We hire 15 annotators with the relevant medical background to work with the annotation process. 
We define 5 intents, 7 actions, and 20 slots and design a set of primer annotation guidelines.
First, each annotator is asked to annotate 5 dialogues and then to report unreasonable, confusing, and ambiguous guidelines with corresponding utterances.
Second, we summarize the confusing issues and improve the guidelines by a high agreement among annotators.
We repeat the above two steps in three rounds and obtain a set of standard annotation guidelines.

\subsection{Human annotation and quality assurance}
To make the annotation more convenient, we build a web-based labeling system similar to \cite{ren2021wizard}, which is available online.\footnote{\url{https://github.com/yanguojun123/Medical-Dialogue}} 
In the system, each annotator is assigned with 5 dialogues each round and is asked to label all utterances following the standard annotation guidelines.
%
To assure annotation quality, we provide:
\begin{enumerate*}[label={(\arabic*)}] 
\item Detailed guidelines.
For each data sample, we introduce the format of the data, the specific labeling task, the examples of various types of labels, and detailed system operations.
\item A real-time feedback paradigm.
We maintain a shared document to track problems and solutions in real-time. 
All annotators can write questions on it; some dialogues with ambiguous labels will be managed: we discussed them with experts and gave the final decision.
\item A semi-automatic quality judgment paradigm.
We adopt a rule-based quality judgment model to assist annotators in re-labeling the untrusted annotations.
\item An entity standardization paradigm.
We use Levinstein distance ratio~\cite{levenshtein1966binary} to compute the similarity between an annotation and an entity in medical knowledge triplet.
If a max similarity score is in [0.9,1], we ask the annotator to replace the annotation with a standard entity from the medical knowledge triplet.
\end{enumerate*}

\subsection{Dataset statistics}
Table~\ref{statistics} shows the data statistics.
\OurResources{} contains 95,408 unlabeled dialogues and 1,557 dialogues with sub-utterance-level \aclp{SL} in the format of intent-slot-value or action-slot-value.
\OurResources{}-large is used for training, and \OurResources{}-base is randomly divided into 657/100/800 dialogues for fine-tuning, validation, testing, respectively.
\OurResources{}-large has 40 domains and 843 diseases.
\OurResources{}-base has 30 domains and 491 diseases.
In \OurResources{}, about 70\% of the dialogues involve multiple services.
\begin{table}[!t]
		\centering
		\setlength{\tabcolsep}{1mm}
        \caption{Statistics of the \OurResources{} dataset.}
		\begin{tabular}{l@{}rrrrr}
		\toprule
			  &  &  \bf Fine- &  &  & \\
			  & \bf Train &  \bf tune & \bf Valid. & \bf Test & \bf Total\\
			\midrule
		    \#Dialogue    & 95,408 & 657 & 100 & 800  & 96,965 \\ 
			\#Utterance & 1,753,624 &10,642 & 1,718 & 13,086 & 1,779,070 \\
			\#Utterance/dialogue     & 18.38  &16.50 & 17.18 & 16.36 & 18.35 \\
			\#Character/dialogue     &  302.29 &311.80 & 332.63 & 302.52 & 316.50 \\
			\#Character/utterance     & 16.18  &19.25 & 19.36 & 19.46  & 16.23 \\
			\#Label/dialogue   & 18.68   &29.85 & 31.01 & 29.85 & 18.86 \\
			\#Label/utterance   &1.0   &1.84 & 1.81  &1.82 & 1.01 \\
		\bottomrule
		\end{tabular}
		
		\label{statistics}
\end{table}



\begin{figure}[!t]
    \centering
    \includegraphics[trim = 0mm 0mm 0mm 0mm, clip, width=0.48\textwidth]{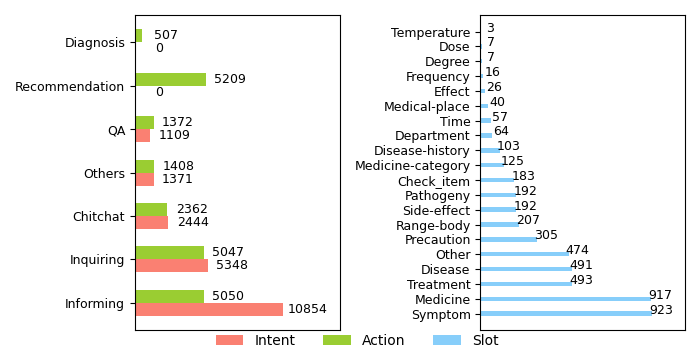}
    \caption{Distribution of utterances  of \OurResources{}-base containing different types of intents/actions (left) and slots (right), respectively.}
    \label{fig:intent_action_slot_dist}
\end{figure}

Figure~\ref{fig:intent_action_slot_dist} shows the number of utterances in \OurResources{}-base distributed over different types of intents/actions and slots. 
In the left chart, there are 5 patient intents (i.e., ``Informing'', ``Inquiring'', ``Chitchat'', ``QA'' and ``Others'') and 7 doctor actions (including 5 intent types plus ``Recommendation'' and ``Diagnosis'').
These cover 25,446 utterances in total, and an utterance might contain multiple intents/actions.
``Informing'' account for the largest proportion (63\%), while ``Diagnosis'' takes up the minimal proportion (2\%). 
It shows that patients have a huge demand for online medical consultations, while doctors are very cautious to make online diagnosis.
The right chart contains 20 types of slots covering 4,825 entities in total.
``Symptom'' (19\%) has the largest proportion of entities, followed by ``Medicine'' (19\%), ``Treatment'' (10\%) and ``Disease'' (10\%). 
In addition, 16\% of label have subordinate value.

\section{The \protect\OurResources{} Benchmarks}

In this section, we unify all tasks as a context-to-text generation task (\S\ref{sec:framework}).
Then we introduce two types of benchmarks, i.e., causal language model (\S\ref{sec:clm}) and conditional causal language model (\S\ref{sec:cclm}). 
Last, we introduce how to enhance models with \ac{CL} to build the state-of-the-art benchmarks (\S\ref{sec:scl}).

\subsection{Unified \ac{MDS} framework}
\label{sec:framework}

We view a \ac{MDS} as a context-to-text generation problem~\cite{hosseini2020simple,pei2020retrospective} and deploy a unified framework called \acs{SeqMDS}.
Formally, given a sequence of dialogue context $X$, a \ac{MDS} aims to generate a system response $Y$ which maximizes the generation probability $P(Y|X)$.
Specifically, all sub-tasks are defined by the following formation.

The \ac{NLU} part of \acs{SeqMDS} aims to generate a list of intent-slot-value triplets $I_t$:
 \begin{equation} 
  I_t=\text{SeqMDS}(C_t),
 \end{equation}
where dialogue history $C_t=[U_1,S_1,\dots,U_t]$ consists of all previous utterances.
And $I_t$ can be used to retrieve a set of related knowledge triplets $K$ from the knowledge base.

The \ac{DPL} part of \acs{SeqMDS} generates the action-slot-value pairs $A_t$ given $C_t$, $I_t$, and $K_t$ as an input:
%
 \begin{equation} 
  A_t=\text{SeqMDS}([C_t;I_t;K_t]).
 \end{equation}
The \ac{NLG} part of \acs{SeqMDS} generates a response based on all previous information:
 \begin{equation} 
  S_t=\text{SeqMDS}([C_t;I_t;K_t;A_t]).
 \end{equation}
\acs{SeqMDS} in the above equations can be implemented by either a causal language model (\S\ref{sec:clm}) or a conditional causal language model (\S\ref{sec:cclm}).
 
\subsection{Causal language model}
\label{sec:clm}
We consider the concatenation $[C;I;K;A;S]$ as a sequence of tokens $X_{1:n}=(x_1,x_2,\ldots,x_n)$.
The $j$-th element $x_j$ can be an intent token (in intent-slot-value pairs), an action token (in action-slot-value pairs), or a general token (in utterances from patients or doctors).
For the $i$-th sequence $X_{1:n}^i$, the goal is to learn the joint probability $p_{\theta}(X_{1:n}^i)$ as:
\begin{equation} 
  p_{\theta}(X_{1:n}^i)=\prod_{j=1}^n (x_j^i|X_{1:j-1}^i).
\end{equation}


\noindent%
The cross-entropy loss is employed to learn parameters $\theta$:
\begin{equation}
 {\mathcal{L}}_{ce}=-\sum_{i=1}^N\sum_{j=1}^{n_i} x^i_j \log p_\theta(x_j^i|X_{1:j-1}^i),
\end{equation}
where $N$ denotes batch size and $n_i$ denotes length of $i$-th utterance.
In this work, we implement the causal language model based on GPT2~\cite{radford2019language}.

\subsection{Conditional casual language model}
\label{sec:cclm}
We consider the concatenation $[C]$, $[C;I;K]$, $[C;I;K;A]$ as the input sequence $X_{1:n}$ and $I$, $A$, $S$ as the generated sequence $Y_{1:m}$ in \ac{NLU}, \ac{DPL} and \ac{NLG}, respectively.

For each input sequence, a transformer encoder is used to convert $X_{1:n}=(x_1,x_2,\ldots,x_n)$ to the corresponding hidden states $H_{1:n}=(h_1,h_1,\ldots,h_n)$,
%
together with the current decoded tokens $Y_{1:j-1}$.
A transformer decoder is used to learn the probability $p_{\theta}(Y_{1:m}|H_{1:n})$ over the vocabulary $V$ at the $j$-th timestamp by:
\begin{equation}
    p_{\theta}(Y_{1:m}|H_{1:n})=\prod_{j=1}^m p_{\theta}(y_j|Y_{1:j-1},H_{1:n}). 
\end{equation}
Similarly, the model can be learned by minimizing the cross-entropy loss as follows:
\begin{equation}
 {\mathcal{L}}_{ce}=-\sum_{i=1}^N\sum_{j=1}^{n_i} y_j^i\log p_{\theta}(y_j^i|Y_{1:j-1}^i,H_{1:n}^i).
\end{equation}
%
In this work, we implement the conditional causal language model based on MT5~\cite{xue2021mT5}.



\subsection{Self-supervised contrastive learning}
\label{sec:scl}
To extend upon the \OurResources{} benchmark approaches introduced so far and enhance model training based on augmented data, we describe a \ac{SCL} approach.
First, we generate data by two heuristic data augmentation approaches, i.e., pseudo labeling (\S\ref{sec:pl}) followed by natural perturbation (\S\ref{sec:np}).
Then, we adopt \acl{CL} (\S\ref{sec:cl}) to assure the models are aware that the augmented data is similar to the original data.

\subsubsection{Pseudo labeling}
\label{sec:pl}
We propose a pseudo labeling algorithm to extend the unlabeled dialogues.
\begin{algorithm}
    \SetAlgoLined
    \SetKwInOut{Input}{\textbf{Input}}\SetKwInOut{Output}{\textbf{Output}} 
    \Input{ 
    $D_L=\{(T_L^i,E_L^i)\}|_{i=1}^{|D_L|}$;~~~$D-D_L$;~~~$R$;
    
    }
    
    \Output{
        $D_P=\{(T_P^i,E_P^i)\}|_{i=1}^{|D_P|}$\;\\
        }
    \BlankLine
    
         \ForEach{$T_P^{i} \in T_P$}
            {
              $\eta, e = MaxSimilariy(T_P^{i},D_L$)
              
              \If{$\eta > \delta$}
                {$E_P^i \longleftarrow e;$}
              \Else 
                {\lForEach{$R^i \in R$}
                     {Update $E_P^i$}
                }            
            }
         
    \SetKwFunction{FMain}{MaxSimilariy}
    \SetKwProg{Fn}{Function}{:}{}
    \Fn{\FMain{$T_P^{i},D_L$}}{
        $\eta = 0; e = null; x = len(T_P^{i}); y = len(T_L^k)$;    
            
        \ForEach{$ T_L^k \in T_L $}
        {
        $\hat{\eta} = 1-LevenshteinDistance(x,y)/(x+y)$;\\
        
        \If{$ \hat{\eta} > \delta $}
            {$ \eta \longleftarrow \hat{\eta}; e \longleftarrow E_L^k $}
        }
        \textbf{return} $ \eta, e; $ 
    }
    
    
    \caption{Pseudo labeling.}
    \label{alg:pseudo_labeling}
\end{algorithm}
As shown in Algorithm~\ref{alg:pseudo_labeling}, 
we decompose the labeled dialogues and unlabeled dialogues into utterance sets $D_L$ and $D_P$, respectively.
Each element of $D_L$ contains an utterance $T_L$ (from user or system) and its corresponding semantic label $E_L$ (in the format of intent-slot-value or action-slot-value).
$R$ is a set of predefined rules, e.g., 
if ``take orally'' is mentioned in some utterance, then the action is  ``Recommendation'' and the slot is ``Medicine''.
The output is $D_P$ with pseudo labels $E_P$.
The main procedure is as follows. 
For each utterance in $D_P$, we calculate the similarities between the current utterance $T_P^{i}$ and all labeled utterances in $D_L$ to get the maximum similarity $\eta$ and the corresponding label $e$.
If $\eta>\delta (\delta=0.8)$, $e$ is assigned as the pseudo label of $T_P^{i}$.
Otherwise, each rule in $R$ is applied to $T_P^{i}$ to update $E_P^i$ gradually.
The similarity is deployed based on Levenshtein distance ~\cite{levenshtein1966binary}, which considers both the overlap rate and the order of characters.

\subsubsection{Natural perturbation}
\label{sec:np}
We use three natural perturbation strategies to extend the labeled dialogues:
\begin{enumerate*}[label={(\arabic*)}] 
\item Alias substitution.
If an utterance contains a drug with an alias, then the drug will be replaced with its alias to obtain a new data.
For example, people from different regions may have different names for the same drug.
\item Back-translation.
Chinese utterances are first translated into English and then back into Chinese to form a new data.
Patients often use colloquial expressions, which motivates us to adopt back-translation to produce formal utterances from the informal ones.
\item Random modification.
We randomly add, delete and replace a character of several medical entities in utterances.
This simulates the common situation: typographical errors in online medical communities.
\end{enumerate*}

\subsubsection{Contrastive Learning}
\label{sec:cl}
We adopt an effective \acl{CL} method to further increase the gains of natural perturbation. 
Following the \ac{CL} framework~\cite{chen20SIMCLR}, we efficiently learn an enhanced representation by contrasting the positive pairs with the negative pairs within a mini-batch.

Given an observed data ($X^i$, $Y^i$), we randomly sample one natural perturbation strategy to get the augmented data ($X^j$, $Y^i$).
Let $z\in \mathbb{R}^d$ denote the sentence representation with $d$ dimension.
We construct the representation of observed and augmented data as a positive pair ($z^i$, $z^j$), and the representation of other data within the mini-batch as negative pairs $\{(z^i, z^{k})\}_{k=1, k \neq i, k \neq  j}^{2N}$.
We compute the pairwise contrastive loss between the observed and augmented data:
\begin{equation}
\begin{split}
    l^{(i,j)} =  -\log \frac{\exp({sim(z^i, {z^j})/\tau)}}{\begin{matrix} \sum_{k=1}^{2N} \mathbbm{1}^{[k\ne i]} \exp(sim({z^i,{z^k})}/\tau) \end{matrix}},\\
    z^i=h_1^i,~~~h_t^i= g(y_{t-1}^i,M^i;\theta),~~~M^i = f(X^i;\theta),
\end{split}
\end{equation}
where $\mathbbm{1}^{[k \neq i]} \in \{0, 1\}$ is an indicator function evaluating to 1 iff  $k \neq i$. 
$\tau$ denotes temperature parameter. 
$f,g$ denote the encoder and decoder respectively and $\theta$ are the model parameters.
The function $sim(u,v)=u^\top v/\|u\| \|v\|$ computes cosine similarity.

For one batch, we minimize contrastive loss across positive pairs, for both ($z^i$, $z^j$) and ($z^j$, $z^i$):
\begin{equation}
    \mathcal{L}_{cl} = \frac{1}{2N} \begin{matrix} \sum_{k=1}^{N} [l(2k-1,2k)+l(2k,2k-1)] \end{matrix}.
\end{equation}

We jointly learn \ac{CL} loss with task-specific cross-entropy loss, and the final loss function is defined as:
\begin{equation}
    \mathcal{L} = \lambda \mathcal{L}_{ce} + (1-\lambda) \mathcal{L}_{cl},
\end{equation}
where $\lambda$ is the coefficient to balance the two training losses.


\section{Evaluating the \protect\OurResources{} benchmarks against the \protect\OurResources{} dataset}

In this section, we first list our evaluation settings, which includes 3 dialogue tasks, 5 benchmark models, 8 automatical metrics and 2 human evaluation metrics. 
Then we report on the results and detailed analysis of the \OurResources{} benchmarks.

\subsection{Tasks}
The \OurResources{} benchmarks  address three tasks, \ac{NLU}, \ac{DPL} and \ac{NLG}:
\begin{description}[leftmargin=\parindent] 
\item[\textbf{\ac{NLU}}] aims to generate a list of intent-slot-value triplets given a dialogue context. 
\item[\textbf{\ac{DPL}}] aims to generate a list of action-slot-value triplets given a dialogue context and a list of intent-slot-value triplets.
\item[\textbf{\ac{NLG}}] aims to generate a response given a dialogue context, intent-slot-value triplets and action-slot-value triplets. 
\end{description}

\subsection{\protect\OurResources{} benchmark models}
\label{sec:models}
We employ several pretrained models as benchmarks:
\begin{description}[leftmargin=\parindent] 
    \item[\textbf{BERT-WWM}] is a BERT~\cite{devlin2018bert} pre-trained on a Chinese Wikipedia corpus~\cite{cui2019pre}.
    \item[\textbf{BERT-MED}]
    is a BERT pre-trained on Chinese medical corpus.\footnote{\url{https://code.ihub.org.cn/projects/1775}} 
    \item[\textbf{GPT2}] is used as a transformer decoder for causal language modeling; we use one pre-trained on Chinese chitchat dialogues~\cite{radford2019language}.\footnote{\url{https://github.com/yangjianxin1/GPT2-chitchat}}
    \item[\textbf{MT5}] is used as a transformer encoder-decoder model for conditional causality modeling. We use the one pre-trained on multilingual C4 dataset~\cite{xue2021mT5}.\footnote{\url{https://github.com/google-research/multilingual-t5}}
    \item[\textbf{MT5+CL}] is an extension of \textbf{MT5} with \acl{CL}.
\end{description}


\subsection{Evaluation setup}
We consider two types of evaluation: automatic (for the \ac{NLU} and \ac{DPL} tasks) and human (for the \ac{NLG} task).
For the \emph{automatic evaluation}, we use 4 metrics to evaluate the \ac{NLU} and \ac{DPL} tasks:
\begin{description}[leftmargin=\parindent]
    \item[\textbf{Micro-F1}] is the intent/action/slot F1 regardless of categories.
    \item[\textbf{Macro-F1}] denotes the weighted average of F1 scores of all categories.
    In this work, we use the proportion of data in each category as the weight.
    \item[\textbf{BLEU}] indicates how similar the generated values of intent/action slots are to the golden ones~\cite{chen2014systematic}.
    \item[\textbf{Combination}] is defined as $0.5*\text{Micro-F1}+0.5*\text{BLEU}$. 
    This measures the overall performance in terms of both intent/action/slot and the generated value.
\end{description}
We use 4 metrics to evaluate the \ac{NLG} task:
\begin{description}[leftmargin=\parindent]
    \item[\textbf{BLEU1} and \textbf{BLEU4}] denote the uni-gram and 4-gram precision, indicating the fraction of the overlapping n-grams out of all n-grams for the responses~\cite{chen2014systematic}. 
    \item[\textbf{ROUGE1}] refers to the uni-grams recall, indicating the fraction of the overlapping uni-grams out of all uni-grams for the responses~\cite{banerjee2005meteor}.
    \item[\textbf{METEOR}] measures the overall performance, i.e., harmonic mean of the uni-gram precision and recall~\cite{lin2004rouge}.
\end{description}

\noindent%
For the \ac{NLG} task, we sample 300 context-response pairs to conduct the \emph{human evaluation}.
We ask annotators to evaluate each response by choosing a score from 0, 1, 2, which denotes bad, neutral, good, respectively.
Each data sample is labeled by 3 annotators.
We define 2 human evaluation metrics: 
\begin{description}[leftmargin=\parindent]
    \item[\textbf{Fluency}] measures to what extent the evaluated responses are fluent.
    \item[\textbf{Specialty}] measures to what extent the evaluated responses provide complete and accurate entities compared with the reference responses.
\end{description}

\subsection{Outcomes}
In this section, we report the results of the \OurResources{} benchmark models (\S\ref{sec:models}) on the NLU, DPL, NLG tasks, respectively. Please note that BERT treats NLU and DPL as a classification task, however, it is inapplicable to NLG task.

\subsubsection{\Acl{NLU}}
Table~\ref{NLU} shows the performance of all models, and the ablation study of MT5 (oracle), on the \ac{NLU} task.
		\label{NLU}
 \begin{table}[!htb]
		\centering
		\setlength{\tabcolsep}{1.5pt}
		\caption{Performance on the \ac{NLU} task. 
		}
		\begin{tabular}{llrccc}
		\toprule
		     \multirow{2}*{} & \multicolumn{2}{c}{Micro-F1/Macro-F1(\%)}          & BLEU(\%) & Combi. \\\cmidrule{2-3}
			                                ~ & Intent        & Intent-Slot    & Value    &  \\
			\midrule
			BERT-WWM                          & 71.76/71.79   & 57.38/58.21    & -        &-\\
			BERT-MED                          & 71.47/71.79   & \textbf{57.64/58.72}      & -        & -\\
			\midrule
			GPT2                              & 73.32/69.23   & 49.23/46.27    & 20.23    & 34.73 \\ 
			\midrule
			MT5                               & 75.32/\textbf{72.67}   & 55.63/\underline{53.07}    & 30.27    & 42.95\\
			-Pseudo labeling                  & 74.33/71.12   & 54.84/52.01    & 30.25    & 42.55\\
		    -Natural perturbation             & 73.90/70.77	  & 53.97/50.99    & 30.68  & 42.33\\     
			-Historical utterances            & 74.43/71.62   & 54.10/51.19    & 29.75   & 41.93\\ 
			\midrule
 			MT5 + CL & \textbf{75.76}/72.65& 55.31/\underline{55.83} & \textbf{30.72} & \textbf{43.02}\\			
		\bottomrule
		\end{tabular}		
\end{table}
First, for intent label identification,
MT5 achieves the best Micro-F1 of 75.32\%, followed by GPT2 of 73.32\%.
MT5 outperforms BERT-WWM/BERT-MED by 3.56\%/3.85\% and GPT2 wins by 1.56\%/1.85\%.
So, MT5 and GPT2 can generate more accurate intent labels compared with BERT models.
Second, for intent-slot label identification, BERT models outperform others by large margins in terms of both Micro-F1 and Macro-F1.
BERT-MED achieves 2.01\%/8.41\% higher Micro-F1 and 5.65\%/12.45\% higher Macro-F1 than MT5 and GPT2.
We believe one of the reasons is that BERT predicts over the label space rather than the whole vocabulary (like GPT2 and MT5), which makes the task easier.
But BERT models are not able to predict the slot-values for the same reason.
Another reason is that unlike intent identification, the training samples of intent-slot identification are inefficient and imbalanced (See Figure~\ref{fig:intent_action_slot_dist}), so the generation models (e.g., MT5 and GPT2) can hardly beat the classification models (e.g., BERT-WWM and BERT-MED).
Third, for value generation, MT5 significantly outperforms GT2 by 10.04\% in terms of BLEU and BERT models are unable to generate values.
It shows that conditional casual language model is more conducive for value generation.
Fourth, MT5 outperforms others in terms of overall performance, i.e., Combination.
We conducted an ablation study, and find that pseudo labeling, natural perturbation, and historical utterances all have positive effect on the overall performance.
Specifically, historical utterances have the largest influence ($-1.02\%$), followed by natural perturbation ($-0.62\%$) and pseudo labeling ($-0.40\%$).
All scores decrease except the BLEU score of MT5 without natural perturbation.
 \begin{CJK*}{UTF8}{gbsn}
This is because that the meaning of entities might be ambiguous after modification, e.g., ``azithromycin'' is replaced by its common name as ``泰力特\ (tylett)'', which is hard to be distinguished from ``力比泰\ (alimta)'' in Chinese.
\end{CJK*}
\Ac{CL} improves the performance of MT5 in terms of most metrics. 
Especially, for \ac{NLU}, it increases 2.76\% of Macro-F1, although it slightly decreases Micro-F1.
\ac{CL} performs better on types of slots that account for a larger proportion of the data (e.g. ``Medicine'' and ``Symptom'' in Figure ~\ref{fig:intent_action_slot_dist}).

\subsubsection{\Acl{DPL}}
Table \ref{DPL} shows the performance of all models, and the ablation study of MT5 (oracle), on the \acs{DPL} task. 
\begin{table}[!htb]
		\centering
		\setlength{\tabcolsep}{1.5pt}
		\caption{Performance on the \acs{DPL} task. The remark ``oracle'' indicates that the ground truth from \acs{NLU} is used instead of the prediction.}
		\label{DPL}
		\begin{tabular}{llrccc}
		\toprule
		     \multirow{2}*{} & \multicolumn{2}{c}{Micro/Macro-F1(\%)}          & BLEU(\%) & Combi. \\\cmidrule{2-3}
			                                ~ & Action        & Action-Slot    & Value    &  \\
			\midrule
			BERT-WWM                          & 52.48/51.98   & 37.23/35.12    & -        &-\\
			BERT-MED                          & 49.83/49.60   & 35.76/34.19    & -        & -\\
			\midrule
			GPT2                              & 43.79/38.80   & 22.37/19.55    & 7.58     & 14.98 \\
			GPT2 (oracle)                     & 45.79/41.63   & 27.22/24.35    & 9.58     & 18.40 \\
			\midrule
			MT5                               & 46.78/41.37   & 26.49/22.58    & 9.41     & 17.95\\
			MT5 (oracle)                      & 53.07/52.07   & 38.58/36.51    & 12.44     & 25.51\\
			-Pseudo labeling                  & 52.04/50.53   & 38.00/36.24    & 11.48     & 24.74\\
		    -Natural perturbation             & 52.40/49.82   & 37.64/35.63    & 12.61    & 25.13\\   
			-Historical utterances            & 50.73/47.97   & 35.98/33.63    & 11.88     & 23.93\\ 
			-External knowledge               & 51.06/48.20   & 31.86/28.74    & 10.90     & 21.38\\	
 			\midrule
 			MT5 (oracle) + CL &\textbf{57.78/55.66}  & \textbf{40.49/38.49}  & \textbf{12.63} & \textbf{26.56}\\
		\bottomrule
		\end{tabular}
		
\end{table}
%
First, MT5 (oracle) outperforms all the other models on all metrics.
Specifically, it outperforms BERT-WWM by 0.59\% and 1.35\% on Micro-F1 for action and action-slot label identification, respectively.
This reveals that MT5 can beat BERT models when more given more information in the input, especially the result from \ac{NLU}.
Besides, it achieves 2.86\% higher BLEU and 7.11\% higher Combination compared with GPT2 (oracle), which indicates that conditional casual language modeling is more effective in this case.
Second, we explore the joint learning performance for MT5 and GPT2, where the prediction from \acs{NLU} is used as an input of \acs{DPL}.
MT5 still outperforms GPT2 by 2.97\% for the Combination performance, specifically 2.99\% for the action label identification, 4.12\% for the action-slot label identification, and 1.83\% for the value generation.
%
Third, we conducted an ablation study on MT5 and find that pseudo labeling, natural perturbation, historical utterances, and external knowledge are still helpful.
Specifically, external knowledge has the largest influence ($-4.13\%$), followed by historical utterances ($-1.58\%$), pseudo labeling ($-0.97\%$) and natural perturbation ($-0.38\%$).
All scores decrease generally.
One exception is that BLEU increases by 0.17\% without natural perturbation.
Similar to the case in \acs{NLU}, some modified entities may cause ambiguity.
\ac{CL} is beneficial  in terms of all evaluation metrics.
Specifically, \ac{CL} increases 1.05\% in Combination, while improving the generation of actions by 4.71\%/3.59\% and action-slots by 1.91\%/1.98\% in terms of Micro/Macro-F1.
Thus \ac{CL} helps the \ac{DPL} task more than it helps the \ac{NLU} task.

\subsubsection{\Acl{NLG}}

Table \ref{tab:nlg} shows the automatic evaluation of GPT2 and MT5, and the ablation study of MT5 (oracle), on \acs{NLG}.
\begin{table}[!htb]
		\centering
		\setlength{\tabcolsep}{3.5pt}
		\caption{Automatic evaluation on the \acs{NLG} task. The remark ``oracle'' indicates that the ground truth from \acs{NLU} and \acs{DPL} is used instead of the prediction.}
		\begin{tabular}{lcccc}
		\toprule
		\multirow{2}*{}
		& \multicolumn{4}{c}{Word-in-Utterance (\%)}
		\\
			                        & BLEU1             & BLEU4             & ROUGE1            & METEOR \\
			 \midrule 
			GPT2                    & 17.29             & 2.11              & 68.15             & 19.55 \\
		    GPT2 (oracle)           & 29.15             & 5.89              & \textbf{74.43}    & 32.62 \\
		    \midrule
			MT5                     & 14.61             & 1.59              & 65.46             & 16.12 \\
			MT5 (oracle)            & 29.71    & 6.92             & 73.49             & 33.11 \\
		    -Pseudo labeling        & 28.76             & 6.59              & 72.89             & 32.41 \\   
			-Natural perturbation   & 26.55             & 6.49              & 71.73             & 30.18 \\     
			-Historical utterances  & 28.21            & 6.62             & 72.69             & 31.98 \\    
			-External knowledge     & 29.66             &  6.97     & 73.41             & 33.06 \\ \midrule
		    
			MT5 (oracle) + CL                            & \textbf{30.70} & \textbf{7.01} & 73.97 & \textbf{33.40}\\
		\bottomrule
		\end{tabular}
		
		\label{tab:nlg}
	\end{table}	
First, MT5 (oracle) outperforms GPT2 (oracle) on METEOR.
Specifically, MT5 (oracle) is 0.56\% and 1.03\% superior on BLEU1 and BLEU4 but 0.93\% inferior on ROUGE1.
It shows that although GPT2 can generate relevant tokens, the generation of MT5 is more precise.
Second, we explore the joint learning performance, where the prediction of \acs{NLU} and \acs{DPL} is used as the input of \acs{NLG}.
We find that MT5 is inferior to GPT2 as METEOR, BLEU1, BLEU4 and ROUGE1 drop by 3.43\%, 2.65\%, 0.52\% and 2.69\%.
It is because that the predictive quality of upstream tasks have more influence on MT5 than GPT2.    
Third, we conduct an ablation study for MT5 (oracle).
We find that pseudo labeling, natural perturbation, historical utterances, and external knowledge are still helpful.
Natural perturbation ($-3.16\%$) is the most influential, followed by historical utterances ($-1.50\%$), pseudo labeling ($-0.95\%$), and external knowledge ($-0.05\%$).
\ac{CL} improves MT5 (oracle) the most on BLUE1 (+0.99\%), followed by ROUGE1 (0.48\%), METEOR (+0.29\%) and BLEU4 (0.09\%).
Unlike the \ac{NLU} and \ac{DPL} tasks, the gains of \ac{CL} for the \ac{NLG} task are limited.

\begin{table}[!htb]
    \centering
    \setlength{\tabcolsep}{21pt}
    \caption{Human evaluation of the \acs{NLG} task. $\kappa$ is the average pairwise Cohen's kappa coefficient between annotators.}
    \begin{tabular}{lcc}
    \toprule
             &  Fluency & Specialty
\\
        \midrule
        GPT2 (oracle)      &  1.72       & 1.04                \\
        MT5 (oracle)    &  1.82      & 1.22                 \\
        \midrule
        Ground truth &  1.91       & 2.00                   \\
        \midrule
        $\kappa$ & 0.64 & 0.62\\ 
    \bottomrule
    \end{tabular}
    
    \label{tab:nlg_human_evaluation}
\end{table}

Table~\ref{tab:nlg_human_evaluation} shows the human evaluation on the \ac{NLG} task.
We did not consider the joint-learned GPT2 and MT5, as they contain the accumulated error from the upstream tasks, which will influence the evaluation of \ac{NLG}.
First, MT5 (oracle) performs better than GPT2 (oracle) on Fluency and Specialty.
This indicates that MT5 can generate more fluent responses that provide more accurate medical knowledge compared with GPT2.
This is consistent with the results of automatic evaluation.
Second, the Fluency score is higher than Specialty for both GPT2 and MT5.
This is because Specialty is more difficult, as generating responses with massive and accurate expertise is more challenging.
Third, the average pairwise Cohen's kappa coefficient is larger than 0.6 for all metrics, which indicates a good annotation agreement.

\subsection{Analysis}
In this section, we analyze one of the strongest performing \OurResources{} benchmarks, MT5, and reflect on the \OurResources{} dataset creation process in terms of dataset size and data acquisition types.

\subsubsection{Dataset size}
\begin{figure}[!htb]
    \centering
    \includegraphics[width=0.49\textwidth]{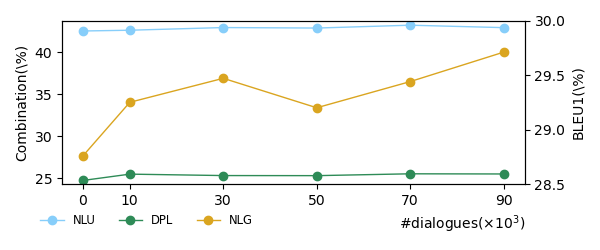}
    \caption{Analysis of data hungry tolerance on the NLU, DPL, and NLG tasks w.r.t. different size of training dialogues.
    The x-axis is the number of training dialogues, and the 0 point denotes no pseudo labeled dialogues.
    The left y-axis is the Combination score on the NLU and DPL tasks, and the right y-axis is the BLEU1 on the NLG task.
    }
    \label{fig:wk}
\end{figure}

The \OurResources{} benchmarks achieve a solid performance for future research to compare against.
Would an increase in the \OurResources{} dataset size have helped to make the benchmarks even more challenging?
To answer this question, we simulate the situation of feeding MT5 with more data by pseudo labeling.
We investigate the performance on the NLU, DPL, NLG tasks with increasing volumes of training dialogues, as shown in Figure~\ref{fig:wk}.
We see that feeding more simulated data has a positive effect on all three tasks, as the overall trends of the lines are upward.
Specifically, NLG is increased by 0.95\% of BLUE1, followed by DPL (+0.79\% of Combination) and NLU(+0.69\% of Combination).
However, the improvement has an upper bound.
For example, adding dialogues to 90K, \ac{NLU} and \ac{DPL} do not improve and even slightly decrease compared with the performance on 70K.
It shows to what extent the current volume of dialogues suffices to approach the upper bound performance.
This helps with the pains-gains trade-off of data acquisition.

\subsubsection{Data acquisition types}
\begin{figure}[!htb]
    \centering
    \includegraphics[width=0.49\textwidth]{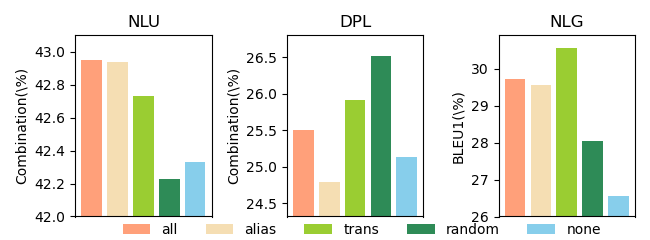}
    \caption{Analysis of data acquisition types on the NLU, DPL, and NLG tasks. 
    We study five settings: (1) all: use all three strategies of natural perturbation; (2) alias: alias substitution; (3) trans: back-translation; (4) random: random modification; (5) none: without any natural perturbations.
    }
    \label{fig:NP}
\end{figure}

What types of data should we expand to enlarge the gains of data acquisition?
As shown in Figure \ref{fig:NP}, we compare the influence of different natural perturbation strategies on all three tasks.
The overall influence of diverse data with lots of perturbation is positive.
The mixture of ``all'' strategies significantly outperforms  the ``none'' of strategies on all tasks.
NLG is developed most by 3.16\%, followed by NLU (+0.62\%) and DPL (+0.38\%).
Besides, different strategies have different influences on different tasks.
The alias strategy improves \ac{NLU} most.
This might be because adding alias entities to data samples helps with entity recognition.
The random strategy has the largest effect on \ac{DPL}, as it can improve the robustness with more input information.
The trans strategy archives the best on \ac{NLG}, as it can generate large-scale dialogues compared with the other two strategies.
Therefore, adding data that can improve the diversity of \OurResources is more than welcomed. 

\section{Conclusion and Future Work}
In this paper, we have introduced key \acfi{ReMeDi}: a dataset and benchmarks.
The \OurResources{} dataset is a multiple-domain, multiple-service dataset with fine-grained labels for \aclp{MDS}.
We focus on providing the community with a new test set for evaluation and provide a small fine-tuning set to encourage low-resource generalization without large, monolithic, labeled training sets.
We consider \ac{NLU}, \ac{DPL} and \ac{NLG} in a unified \acs{SeqMDS} framework, based on which, we deploy several state-of-the-art pretrained language models, with contrastive learning, as benchmarks, the \OurResources{} benchmarks.
We have also evaluated the \OurResources{} benchmarks against the \OurResources{} dataset. 
Both the \OurResources{} datasets and benchmarks are available online; please see the Appendix for details.

The resources released with this work have broader implications in that: (1) The fine-gained labels provided with \OurResources{} can help research on the interpretability of \aclp{MDS}. (2) The performance of the baseline models are far from satisfactory; therefore, we hope that the \OurResources{} resources facilitate and encourage research in low-resource \aclp{MDS}.

One limitation of the \OurResources{} dataset is that we do not provide explicit boundaries between dialogue sessions with different service types. 
This makes it challenging to explicitly model relationships among multiple services.

As to future work, 
on the one hand, we will extend \OurResources{} with service boundary labels to facilitate research on dialogue context modeling among multiple services.
On the other hand, we will extend \OurResources{} with more languages to help study multilingual \acp{MDS}.
Last but not least, we call for studies to improve the benchmark performance, as well as conduct underexplored research, e.g., dialogue tasks for rare diseases under extremely low-resource settings.

\appendix
\section{Appendix}

\subsection{Resources}
All resources presented in this paper, the dataset, code for the baselines, and evaluation scripts are shared at \url{https://github.com/yanguojun123/Medical-Dialogue}.
The shared resources are organized in multiple folders:
\begin{enumerate*}[label={(\arabic*)}]
\item The folder ``annotation\_guide'' contains the code and guidelines for the labeling system. 
\item The folder ``data'' contains the ReMeDi-large and ReMeDi-base datasets.
Each dialogue consists of multiple turns of utterances from doctors or patients, identified with a unique dialogue id. 
Each utterance in a dialogue turn is provided with: turn id, dialogue role, utterance text, and label list. 
Each label consists of sub-sentence text, the start/end position of sub-sentences, and the intent-slot-value or action-slot-value labels.
\item The folder ``data\_process'' contains the code for processing the crawled raw data, the pseudo labeling and natural perturbation. 
\item The folder ``model'' contains the code of the benchmark models based on BERT, GPT2, and MT5.
\item The folder ``evaluate'' contains the code for automatic evaluation in terms of all metrics.
\end{enumerate*}
All resources are licensed under the MIT license.





\subsection{Implementation details}
BERT-WWM and BERT-MED use 12 transformer blocks with 12 attention heads and the hidden size is 768.
The maximum length of input tokens is 512 and the learning rate is 2e-5.

GPT2 uses 10 transformer decoder blocks with 12 attention heads and the hidden size is 768.
MT5 uses 8 transformer encoder blocks followed by 8 decoder blocks with 12 attention heads and the hidden size is 512.
For GPT2 and MT5, the maximum length of input tokens is 800 and the learning rate is 1.5e-4.
We set the coefficient $\lambda$ as 0.8 and the temperature $\tau$ as 0.5 for the contrastive learning. 

We fine-tune the models on three training sets produced by pseudo labeling, natural perturbation, and human annotation, respectively. 
We use AdamW~\cite{adam_optimizer} as the optimization algorithm.
The maximum training epochs are set to 30.
We implemented benchmark models by PyTorch~\cite{paszke2019pytorch}.
Our model is trained with 4 Nvidia TITAN RTX GPUs with 20 GB of memory.
The results reported in this work can be reproduced with the random seed fixed.

\clearpage
\bibliographystyle{ACM-Reference-Format}
\balance
\bibliography{references}

\end{document}